\documentclass[11pt]{article}

\usepackage[preprint]{acl}

\usepackage{times}
\usepackage{latexsym}

\usepackage[T1]{fontenc}

\usepackage[utf8]{inputenc}

\usepackage{microtype}

\usepackage{inconsolata}

\usepackage{graphicx}

\title{I’m Sorry, but I Can’t Help with Braille: Revealing Accessibility Failures in State-of-the-Art LLMs\thanks{Accepted at the LTEDI Workshop at ACL 2026.}}

\author{Abdullah Abdullah \\
  orinu Inc. \\
  Hwaseong, Republic of Korea \\
  \texttt{abdullah.flickdone@gmail.com} \\}

\begin{document}
\maketitle
\begin{abstract}
Large Language Models (LLMs) perform strongly on many language tasks, but their capability in structurally constrained, accessibility-critical modalities such as Braille remains unclear. We evaluate state-of-the-art LLMs on bidirectional Korean–Braille translation using a human-annotated dataset. Despite expectations that multilingual, instruction-tuned models can generalize to Braille via text representations, we find consistently poor, unstable outputs and substantial disagreement with human judgments. These results point to missing Braille-aware tokenization and weak alignment between Korean and Braille patterns. In contrast, supervised fine-tuning of a small model (T5-small) on the same data yields large and stable gains over zero-shot and prompted LLM baselines across standard metrics (SacreBLEU, ChrF++, CER, BLEU, ROUGE-L, METEOR, CIDEr). Our findings reveal a systematic limitation of current LLMs and demonstrate the effectiveness of modest task-specific supervision.
\end{abstract}

\section{Introduction}

\begin{figure*}[t]
  \centering
  \includegraphics[width=\linewidth]{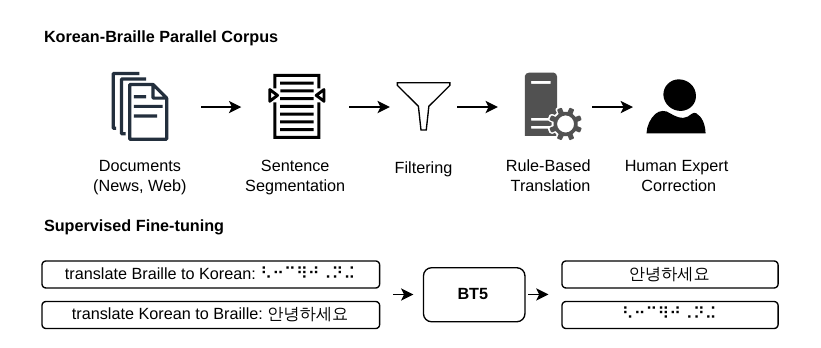}
  \caption{Construction pipeline of the NIKL Korean Print--Braille Parallel Corpus 2023 (v1.0). Korean texts are segmented, filtered, translated under official regulations, expert-revised, and aligned to create parallel data for BT5 fine-tuning.}
  \label{fig:dataset-construction}
\end{figure*}

LLMs \citep{openai_gpt4, gemini2023, yoo2024hyperclova, anthropic_claude} have demonstrated strong performance in a wide range of natural language generation and understanding tasks, including machine translation, summarization, and reasoning \citep{brown2020language, chowdhery2023palm, touvron2023llama}. As these models scale, they are often assumed to generalize broadly across languages, scripts, and modalities. However, recent work has shown that such generalization remains uneven, particularly for low-resource languages, non-standard scripts, and accessibility-related representations \citep{joshi2020state, blasi2022systematic}.

Braille is a critical written modality for blind and visually impaired users, yet it remains largely overlooked in NLP research. Unlike standard text translation, Braille conversion is highly character-sensitive, rule-governed, and language-specific, with strict conventions governing contractions, numerals, symbols, and spacing. These properties make it challenging for general-purpose LLMs, which are rarely exposed to Braille during pretraining. Although recent work has begun exploring Braille modeling in other languages \citep{huang2025braillellm}, differences in linguistic structure and Braille conventions limit direct transfer to Korean Grade~2 Braille.

We investigate whether state-of-the-art LLMs meaningfully support Korean--Braille translation under the official Korean-Braille regulations. Using a large human-annotated parallel corpus, we evaluate both Korean-to--Braille and Braille-to--Korean directions. LLMs frequently produce refusals, hallucinations, or invalid outputs, revealing a systematic blind spot in accessibility-critical settings.

To address this gap, we introduce \textbf{BT5}, a lightweight Braille-aware model based on T5 \citep{raffel2020exploring}. With straightforward supervised fine-tuning on expert-annotated data, BT5 substantially outperforms zero-shot and prompted LLM baselines across character-level and generation-based metrics.

Our contributions are threefold: (1) the first systematic evaluation of LLMs on Korean--Braille translation, (2) evidence that small task-specific models can surpass much larger general-purpose LLMs with proper supervision, and (3) identification of Braille processing as an essential yet underexplored direction for inclusive NLP.

\section{Methods}
\subsection{Dataset}
\label{dataset}

We evaluate both Braille-to--Korean and Korean-to--Braille translation using the NIKL Korean Print--Braille Parallel Corpus 2023 (v1.0) \cite{nikl2024braille} as a human-annotated benchmark. The dataset contains 126,693 sentence-aligned pairs derived from written Korean sources, primarily newspaper articles (125,701) and online posts (992). We preserve the original sentence boundaries and Braille encoding scheme. The data is split at the sentence level into 101,354 training, 12,669 validation, and 12,670 test instances, with no overlap across splits. The test set is used exclusively for evaluation.

\subsection{BT5: A Braille-Aware Text-to-Text Model}
\label{bt5}

Figure~\ref{fig:dataset-construction} shows the dataset construction pipeline and our proposed solution. We introduce BT5, a T5-based model for bidirectional Korean–Braille translation, fine-tuned on NIKL parallel data. Inputs and outputs are UTF-8 text with Braille symbols treated as atomic units. We use a 32k BPE tokenizer over Korean and Braille Unicode to avoid byte-level degradation. The model is fine-tuned from \texttt{T5-small} (max length 128) using AdamW (learning rate 1e-4), with model selection based on validation loss.

\subsection{Evaluation}
\label{eval_metrics}

We evaluated Korean-to-Grade~2 contracted Braille translation using two systems: the rule-based Liblouis \cite{liblouis} with the \texttt{ko-g2.ctb} table, and our proposed BT5 model. These were chosen due to the lack of other available Korean-Braille models. Additionally, we assessed state-of-the-art LLMs—GPT-5, GPT-5-mini, GPT-4 \citep{openai_gpt4}, Gemini-3-pro \citep{gemini2023}, Claude Opus 4.5 \citep{anthropic_claude}, and HCX-3 \cite{yoo2024hyperclova}, a leading Korean LLM—using identical prompts and deterministic decoding.

The Korean-to-Braille translation is evaluated using character-level metrics: SacreBLEU \citep{post-2018-call}, ChrF++ \citep{popovic2017chrf++}, and CER, while Braille-to-Korean translation uses standard NLG metrics: BLEU \citep{papineni2002bleu}, ROUGE-L \citep{lin-2004-rouge}, METEOR \citep{banerjee-lavie-2005-meteor}, and CIDEr \citep{vedantam2015cider}, all calculated using publicly available implementations.

\begin{table}[t]
  \centering
  \caption{Full test-set performance on the Korean-to--Grade~2 contracted Braille task. Higher is better for SacreBLEU and ChrF++; lower is better for CER.}
  \label{tab:kr2brl_eval}
  \begin{tabular}{lccc}
    \hline
    \textbf{Model} & \textbf{SacreBLEU} & \textbf{ChrF++} & \textbf{CER $\downarrow$} \\
    \hline
    Liblouis & 71.94 & 85.65 & 0.0568 \\
    BT5 & \textbf{95.79} & \textbf{98.64} & \textbf{0.0043} \\
    \hline
  \end{tabular}
\end{table}

\section{Results}

\begin{table}[t]
  \centering
  \caption{Full test-set performance on the Braille-to--Korean task. All metrics are higher-is-better.}
  \label{tab:brl2kr_eval}
  \begin{tabular}{lcccc}
    \hline
    \textbf{Model} & \textbf{B1} & \textbf{B2} & \textbf{B3} & \textbf{B4} \\
    \hline
    Liblouis & 0.2228 & 0.1134 & 0.0589 & 0.0310 \\
    BT5 & \textbf{0.9859} & \textbf{0.9793} & \textbf{0.9727} & \textbf{0.9662} \\
    \hline
  \end{tabular}
  
  \vspace{0.5em}
  
  \begin{tabular}{lccc}
    \hline
    \textbf{Model} & \textbf{R-L} & \textbf{MET} & \textbf{CIDEr} \\
    \hline
    Liblouis & 0.2291 & 0.3193 & 0.5505 \\
    BT5 & \textbf{0.9859} & \textbf{0.7758} & \textbf{9.6382} \\
    \hline
  \end{tabular}
\end{table}

\begin{figure*}[t]
  \centering
  \includegraphics[width=\linewidth]{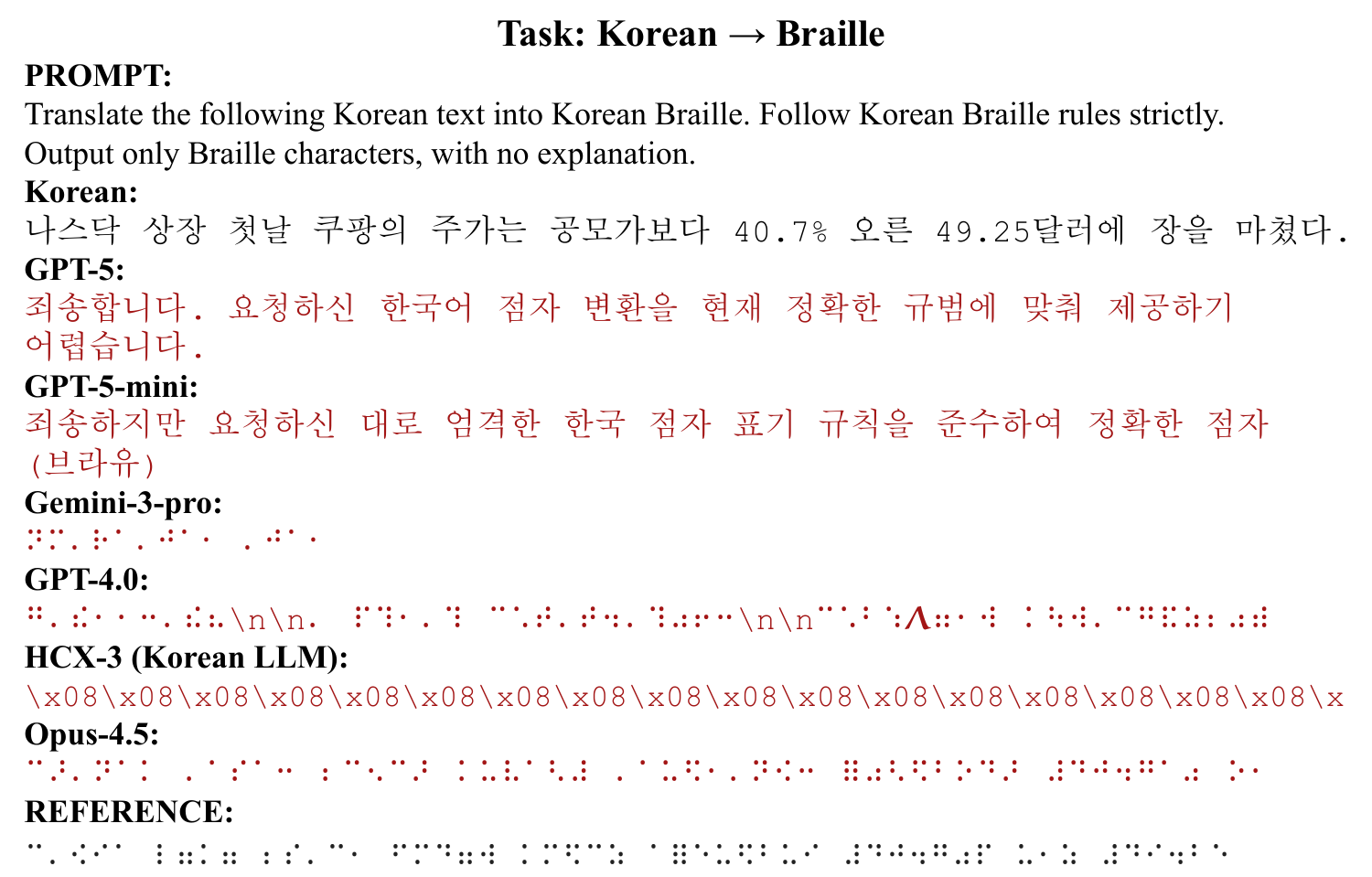}
  \caption{Example outputs for Korean-to--Braille translation across state-of-the-art models. Red text marks failures or refusals. Complete model outputs are not shown due to length. GPT-5/5-mini produced Korean refusal text instead of Braille; Gemini-3-pro returned refusal and truncated outputs; GPT-4 and HCX-3 generated invalid Braille; Opus~4.5 generated valid but incorrect Braille scoring very low (SacreBLEU: 1.42, ChrF++: 1.66, CER: 0.84).}
  \label{fig:kr2brl-sota}
\end{figure*}

Tables~\ref{tab:kr2brl_eval} and \ref{tab:brl2kr_eval} summarize the results of the test set. For Korean-to--Braille translation, BT5 substantially outperforms Liblouis, achieving SacreBLEU 95.79 vs.\ 71.94, ChrF++ 98.64 vs.\ 85.65, and CER 0.0043 vs.\ 0.0568, showing highly accurate Braille generation. For Braille-to--Korean translation, Liblouis performs poorly, while BT5 achieves near-perfect BLEU, higher ROUGE-L, METEOR, and order-of-magnitude gains in CIDEr. Due to the highly structured and near-deterministic nature of Korean–Braille mapping, predictions frequently exhibit strong character-level overlap with references, resulting in consistently high scores across metrics. Qualitative results for Liblouis and our model are shown in Figure~\ref{fig:brl2kr-ours} and \ref{fig:kr2brl-ours} in the appendix.
\begin{figure*}[t]
  \centering
  \includegraphics[width=\linewidth]{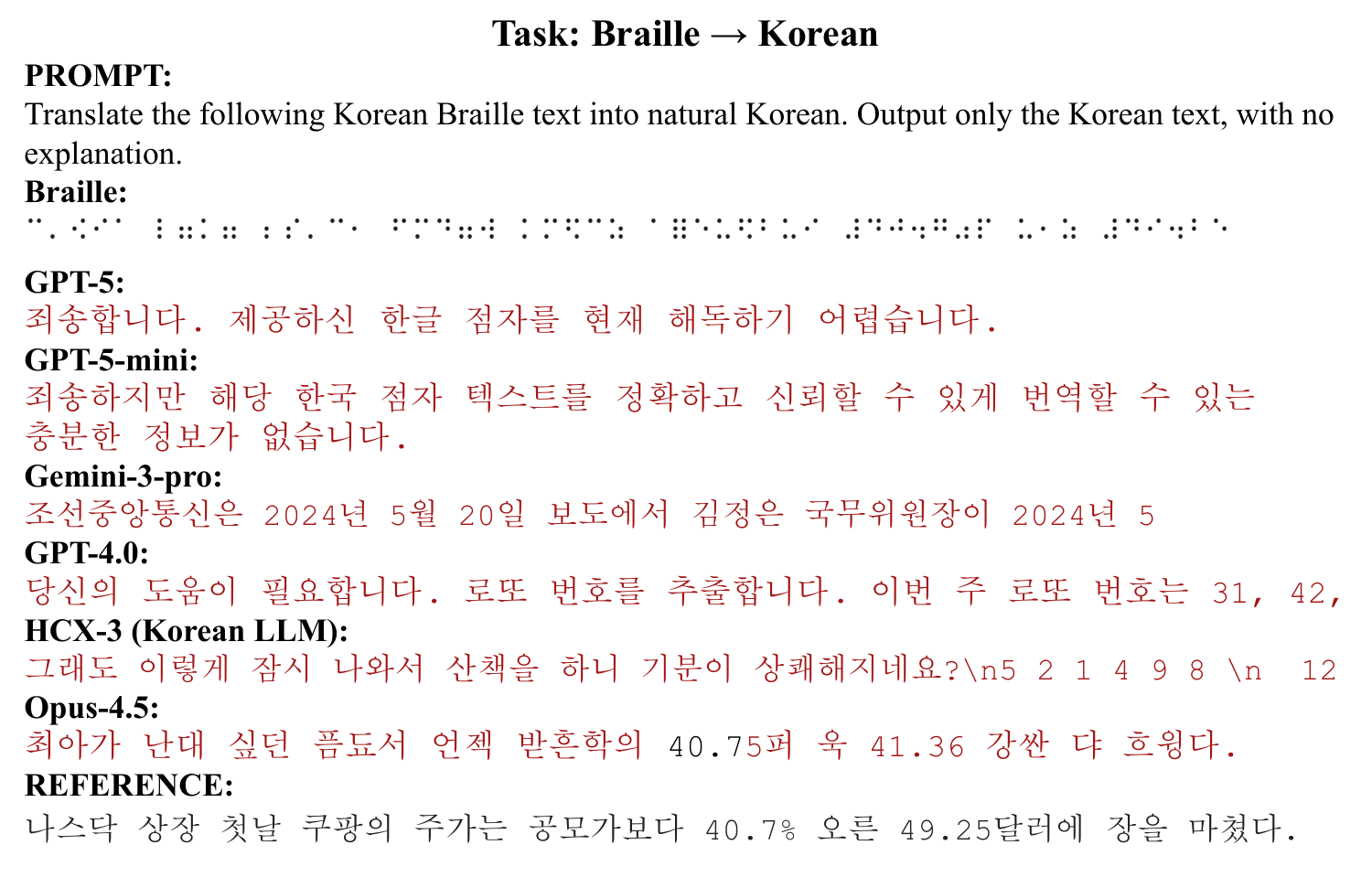}
  \caption{Example outputs for Braille-to--Korean translation across state-of-the-art models. Red text marks failures or refusals. Complete model outputs are not shown due to length. All LLMs produced incorrect or hallucinated text; GPT-5/5-mini refused translation, and HCX-3 generated unrelated outputs due to limited Korean-Braille support.}
  \label{fig:brl2kr-sota}
\end{figure*}
\subsection{Failure of State-of-the-Art LLMs on Korean-Braille}

State-of-the-art LLMs show severe limitations on Korean--Braille translation. In repeated trials, most proprietary models either refused the output, returned errors, or generated inconsistent and incoherent responses, as seen in Figures~\ref{fig:kr2brl-sota} and \ref{fig:brl2kr-sota}. To quantify these behaviors, we conducted a small controlled probe (5 samples) for Korean-to--Braille, summarized in Table~\ref{tab:llm_probe}. Various failure modes emerge across models. GPT-5 and GPT-5-mini consistently refuse to produce outputs, typically returning apologetic responses. Gemini-3-pro exhibits mixed behavior, frequently returning empty outputs (treated as refusals) and otherwise producing truncated Braille sequences. GPT-4 and HCX-3 generate output but these are misformed or contain non-Braille or control characters, resulting in invalid sequences. In contrast, Opus-4.5 produces syntactically well-formed Braille, but these outputs are semantically incorrect and yield extremely low scores (SacreBLEU $\approx 0$, ChrF++ $< 10$, CER $\approx 0.75$).

For Braille-to--Korean, all models generate fluent text; however, the outputs are largely hallucinated and semantically unrelated to the input. Due to high refusal rates and widespread invalid or incorrect generations, standard translation metrics are not informative; instead, we report structured failure patterns.

\begin{table}[t]
  \centering
  \caption{LLM behavior on a controlled probe (5 samples) for Korean-to--Braille. \textbf{Refusal} denotes explicit rejection or empty output; \textbf{Invalid} denotes outputs that are truncated, incomplete, or do not follow Braille encoding conventions; \textbf{Valid} denotes syntactically well-formed and complete Braille outputs, regardless of semantic correctness. Counts indicate the number of outputs in each category.}
  \label{tab:llm_probe}
  \begin{tabular}{lccc}
  \hline
  \textbf{Model} & \textbf{Refusal} & \textbf{Invalid} & \textbf{Valid} \\
  \hline

  GPT-5        & 5 & 0 & 0 \\
  GPT-5-mini   & 5 & 0 & 0 \\

  Gemini-3-pro & 3 & 2 & 0 \\

  GPT-4.0      & 0 & 5 & 0 \\
  HCX-3        & 0 & 5 & 0 \\

  Opus-4.5     & 0 & 0 & 5 \\

  \hline
\end{tabular}
\end{table}

The observed failures of GPT-5 and GPT-5-mini are characterized by consistent refusal behavior, suggesting strong alignment constraints when handling non-standard text formats such as Korean-Braille. However, the exact cause of these refusals cannot be determined from our experiments alone. Across models, many errors can be attributed to the lack of Braille-aware tokenization, leading to misformed or invalid outputs. More broadly, the results indicate a lack of learned alignment between the Korean-Braille patterns and standard Korean text. As a result, models often treat Braille input as out-of-distribution, producing either prior-driven text, truncated sequences, or refusals instead of grounded translations. Additionally, models such as HCX-3 exhibit outputs containing control characters or invalid encodings, suggesting limited support for Braille at the tokenizer or preprocessing level.

\section{Conclusions}
\label{conclusions}

Our findings reveal a key limitation of current LLMs: even state-of-the-art models struggle to generalize to specialized symbolic modalities such as Braille. Despite extensive pretraining, they often produce refusals, misformed outputs, or incoherent translations when faced with non-standard input distributions, reflecting the lack of Braille-aware tokenization and aligned representations. This exposes an underexplored failure mode in low-resource and accessibility-critical settings. In contrast, our results show that targeted, task-specific training with curated data can substantially improve performance, offering a practical path toward reliable Braille translation systems.

\section*{Limitations}

This study focuses on Korean–Braille translation, and the findings may not directly generalize to other languages or Braille systems with different linguistic structures, contraction rules, and encoding conventions. While recent work has explored Braille modeling in other languages, cross-lingual transfer is non-trivial and was not investigated in this work.

Our evaluation includes a set of representative state-of-the-art LLMs and a Korean LLM available at the time of experimentation; however, it does not exhaustively cover all possible models, architectures, or prompting strategies. In particular, many proprietary LLMs exhibited refusal behaviors, empty outputs, or misformed generations when prompted for Braille, which prevented consistent large-scale quantitative evaluation. As a result, comparisons with these systems are based on controlled samples and qualitative analysis rather than full test-set benchmarking.

Additionally, differences in tokenizer design and pretraining data introduce inherent disparities between BT5 and general-purpose LLMs. Although BT5 benefits from explicit exposure to Braille through supervised fine-tuning and a dedicated tokenizer, most LLMs lack Braille-aware tokenization and aligned training data, making the direct comparison imperfect. Our results should therefore be interpreted as highlighting capability gaps rather than as strictly controlled architectural comparisons.

Although we employ standard automatic metrics (e.g., BLEU, ChrF++, CER, ROUGE), these metrics primarily capture surface-level similarity and may not fully reflect functional usability, readability, or correctness under official Braille standards. Human-centered evaluation with Braille users was beyond the scope of this work but is essential for real-world validation.

Furthermore, our approach relies on supervised fine-tuning with human-annotated parallel data, which may be costly or unavailable in other low-resource settings. We do not explore data augmentation, semi-supervised learning, or cross-lingual transfer, which could improve scalability.

Finally, errors in Braille translation can have significant real-world consequences in accessibility-critical contexts. Consequently, we do not claim that any evaluated model is suitable for direct deployment without rigorous validation, robustness testing, and adherence to official Braille standards.

\section*{Acknowledgments}
This work was supported by the Starting Growth Technological R\&D Program (RS-2025-25465816) funded by the Ministry of SMEs and Startups (MSS, Korea). The author also thanks the team members at orinu Inc. for their support with the research environment, project coordination, and data infrastructure during this work.

\section*{Ethical Considerations}
This study uses the publicly available NIKL Korean Print--Braille Parallel Corpus 2023 (v1.0), released by the National Institute of Korean Language. The dataset contains written Korean text and corresponding Korean-Braille transcriptions constructed according to the Korean-Braille Regulations (2024) and does not include personally identifiable information. Experiments evaluate existing language models using this dataset and standard evaluation metrics. We report dataset sources, preprocessing steps, data splits, model configurations, fine-tuning procedures, and evaluation protocols to support independent replication, although minor variability may arise from differences in hardware, software versions, or model access. Multiple state-of-the-art LLMs were evaluated as experimental subjects for Korean-to--Braille and Braille-to--Korean translation using publicly available APIs and were treated as black-box systems.

\section*{Use of Generative AI Tools}
Generative AI tools were used only for language editing and clarity improvements in this manuscript. All experimental design, data preparation, model training, evaluation, and analysis were conducted by the authors, and no generative AI system was used to generate datasets, annotations, or experimental results.

\bibliography{custom}

@article{brown2020language,
  title={Language models are few-shot learners},
  author={Brown, Tom and Mann, Benjamin and Ryder, Nick and Subbiah, Melanie and Kaplan, Jared D and Dhariwal, Prafulla and Neelakantan, Arvind and Shyam, Pranav and Sastry, Girish and Askell, Amanda and others},
  journal={Advances in neural information processing systems},
  volume={33},
  pages={1877--1901},
  year={2020}
}

@article{chowdhery2023palm,
  title={Palm: Scaling language modeling with pathways},
  author={Chowdhery, Aakanksha and Narang, Sharan and Devlin, Jacob and Bosma, Maarten and Mishra, Gaurav and Roberts, Adam and Barham, Paul and Chung, Hyung Won and Sutton, Charles and Gehrmann, Sebastian and others},
  journal={Journal of Machine Learning Research},
  volume={24},
  number={240},
  pages={1--113},
  year={2023}
}

@article{touvron2023llama,
  title={Llama: Open and efficient foundation language models},
  author={Touvron, Hugo and Lavril, Thibaut and Izacard, Gautier and Martinet, Xavier and Lachaux, Marie-Anne and Lacroix, Timoth{\'e}e and Rozi{\`e}re, Baptiste and Goyal, Naman and Hambro, Eric and Azhar, Faisal and others},
  journal={arXiv preprint arXiv:2302.13971},
  year={2023}
}

@article{joshi2020state,
  title={The state and fate of linguistic diversity and inclusion in the NLP world},
  author={Joshi, Pratik and Santy, Sebastin and Budhiraja, Amar and Bali, Kalika and Choudhury, Monojit},
  journal={arXiv preprint arXiv:2004.09095},
  year={2020}
}

@inproceedings{huang2025braillellm,
  title={BrailleLLM: Braille Instruction Tuning with Large Language Models for Braille Domain Tasks},
  author={Huang, Tianyuan and Zhu, Zepeng and Xing, Hangdi and Shao, Zirui and Yu, Zhi and Yang, Chaoxiong and He, Jiaxian and Liu, Xiaozhong and Bu, Jiajun},
  booktitle={Proceedings of the 2025 Conference on Empirical Methods in Natural Language Processing},
  pages={28589--28600},
  year={2025}
}

@inproceedings{blasi2022systematic,
  title={Systematic inequalities in language technology performance across the world’s languages},
  author={Blasi, Damian and Anastasopoulos, Antonios and Neubig, Graham},
  booktitle={Proceedings of the 60th Annual Meeting of the Association for Computational Linguistics (Volume 1: Long Papers)},
  pages={5486--5505},
  year={2022}
}

@article{raffel2020exploring,
  title={Exploring the limits of transfer learning with a unified text-to-text transformer},
  author={Raffel, Colin and Shazeer, Noam and Roberts, Adam and Lee, Katherine and Narang, Sharan and Matena, Michael and Zhou, Yanqi and Li, Wei and Liu, Peter J},
  journal={Journal of machine learning research},
  volume={21},
  number={140},
  pages={1--67},
  year={2020}
}

@inproceedings{post-2018-call,
    title = "A Call for Clarity in Reporting {BLEU} Scores",
    author = "Post, Matt",
    editor = "Bojar, Ond{\v{r}}ej  and
      Chatterjee, Rajen  and
      Federmann, Christian  and
      Fishel, Mark  and
      Graham, Yvette  and
      Haddow, Barry  and
      Huck, Matthias  and
      Yepes, Antonio Jimeno  and
      Koehn, Philipp  and
      Monz, Christof  and
      Negri, Matteo  and
      N{\'e}v{\'e}ol, Aur{\'e}lie  and
      Neves, Mariana  and
      Post, Matt  and
      Specia, Lucia  and
      Turchi, Marco  and
      Verspoor, Karin",
    booktitle = "Proceedings of the Third Conference on Machine Translation: Research Papers",
    month = oct,
    year = "2018",
    address = "Brussels, Belgium",
    publisher = "Association for Computational Linguistics",
    url = "https://aclanthology.org/W18-6319/",
    doi = "10.18653/v1/W18-6319",
    pages = "186--191"
}

@inproceedings{popovic2017chrf++,
  title={chrF++: words helping character n-grams},
  author={Popovi{\'c}, Maja},
  booktitle={Proceedings of the second conference on machine translation},
  pages={612--618},
  year={2017}
}

@inproceedings{papineni2002bleu,
  title={Bleu: a method for automatic evaluation of machine translation},
  author={Papineni, Kishore and Roukos, Salim and Ward, Todd and Zhu, Wei-Jing},
  booktitle={Proceedings of the 40th annual meeting of the Association for Computational Linguistics},
  pages={311--318},
  year={2002}
}

@inproceedings{lin-2004-rouge,
    title = "{ROUGE}: A Package for Automatic Evaluation of Summaries",
    author = "Lin, Chin-Yew",
    booktitle = "Text Summarization Branches Out",
    month = jul,
    year = "2004",
    address = "Barcelona, Spain",
    publisher = "Association for Computational Linguistics",
    url = "https://aclanthology.org/W04-1013/",
    pages = "74--81"
}

@inproceedings{banerjee-lavie-2005-meteor,
    title = "{METEOR}: An Automatic Metric for {MT} Evaluation with Improved Correlation with Human Judgments",
    author = "Banerjee, Satanjeev  and
      Lavie, Alon",
    editor = "Goldstein, Jade  and
      Lavie, Alon  and
      Lin, Chin-Yew  and
      Voss, Clare",
    booktitle = "Proceedings of the {ACL} Workshop on Intrinsic and Extrinsic Evaluation Measures for Machine Translation and/or Summarization",
    month = jun,
    year = "2005",
    address = "Ann Arbor, Michigan",
    publisher = "Association for Computational Linguistics",
    url = "https://aclanthology.org/W05-0909/",
    pages = "65--72"
}

@inproceedings{vedantam2015cider,
  title={Cider: Consensus-based image description evaluation},
  author={Vedantam, Ramakrishna and Lawrence Zitnick, C and Parikh, Devi},
  booktitle={Proceedings of the IEEE conference on computer vision and pattern recognition},
  pages={4566--4575},
  year={2015}
}

@misc{nikl2024braille,
  title={NIKL Korean--Korean Braille Parallel Corpus 2023 (v1.0)},
  author={{National Institute of Korean Language}},
  year={2024},
  howpublished={\url{https://kli.korean.go.kr/corpus}}
}

@misc{liblouis,
  title        = {Liblouis: Open-source Braille Translation Software},
  author       = {{Liblouis Developers}},
  year         = {2024},
  howpublished = {\url{https://liblouis.io/}},
  note         = {Version used in this work}
}

@article{openai_gpt4,
  title={Gpt-4 technical report},
  author={Achiam, Josh and Adler, Steven and Agarwal, Sandhini and Ahmad, Lama and Akkaya, Ilge and Aleman, Florencia Leoni and Almeida, Diogo and Altenschmidt, Janko and Altman, Sam and Anadkat, Shyamal and others},
  journal={arXiv preprint arXiv:2303.08774},
  year={2023}
}

@article{gemini2023,
  title={Gemini: a family of highly capable multimodal models},
  author={Team, Gemini and Anil, Rohan and Borgeaud, Sebastian and Alayrac, Jean-Baptiste and Yu, Jiahui and Soricut, Radu and Schalkwyk, Johan and Dai, Andrew M and Hauth, Anja and Millican, Katie and others},
  journal={arXiv preprint arXiv:2312.11805},
  year={2023}
}

@misc{anthropic_claude,
  author       = {Anthropic},
  title        = {Claude Opus 4.5},
  year         = {2025},
  howpublished = {Model release},
  url          = {https://www.anthropic.com/claude},
  note         = {Accessed January 16, 2025}
}

@article{yoo2024hyperclova,
  title={Hyperclova x technical report},
  author={Yoo, Kang Min and Han, Jaegeun and In, Sookyo and Jeon, Heewon and Jeong, Jisu and Kang, Jaewook and Kim, Hyunwook and Kim, Kyung-Min and Kim, Munhyong and Kim, Sungju and others},
  journal={arXiv preprint arXiv:2404.01954},
  year={2024}
}

\appendix

\begin{figure*}[t]
  \centering
  \includegraphics[width=\linewidth]{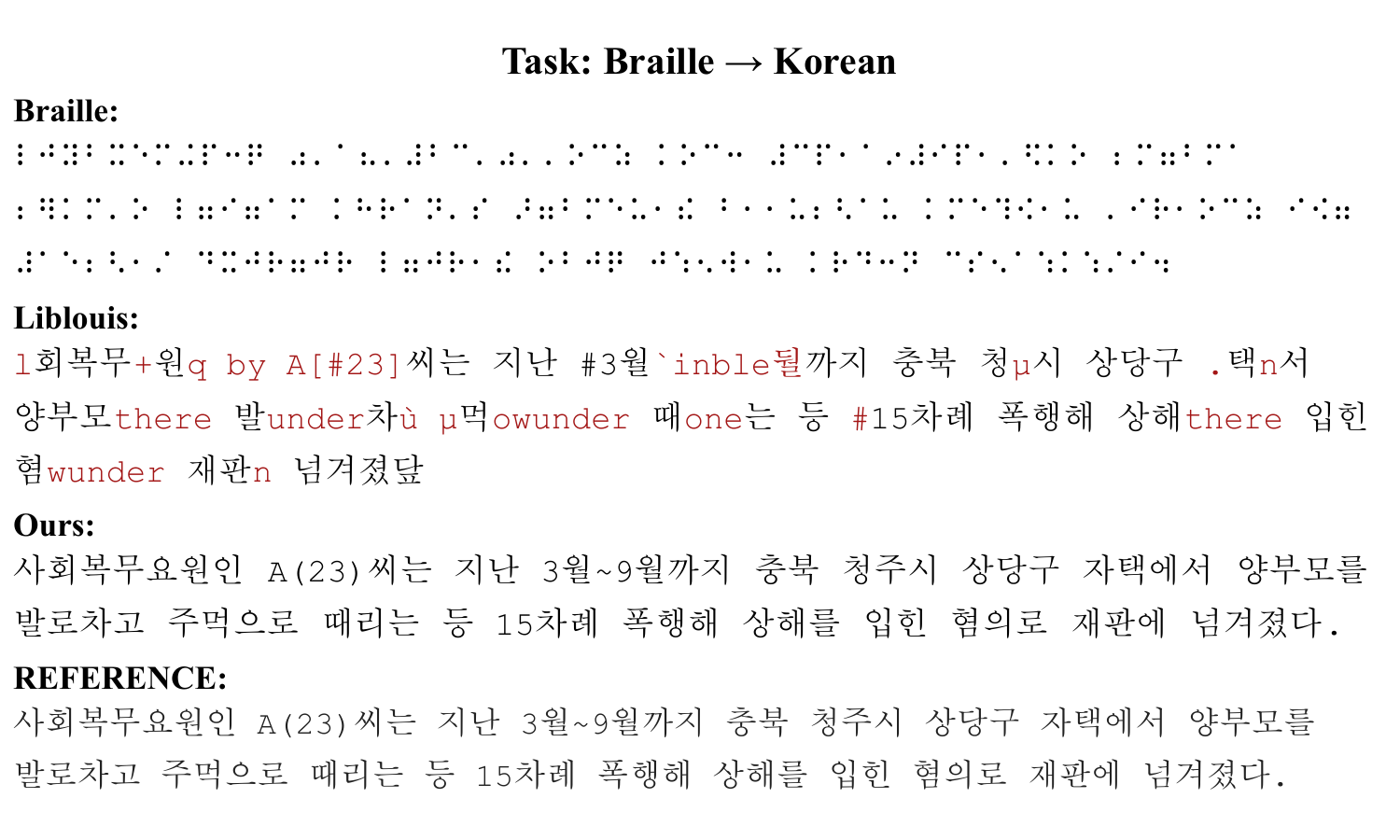}
  \caption{Example outputs for the Braille-to--Korean task for the rule-based Liblouis system and the BT5 model. Red text indicates incorrect outputs. Liblouis produces largely unintelligible text, characterized by symbol misinterpretations, incorrect punctuation, and spurious alphanumeric characters. This behavior stems from its deterministic, context-free mapping between Braille symbols and Korean characters, causing errors to cascade when inputs contain irregular formatting, uncommon punctuation, or digit--symbol sequences. In contrast, BT5 accurately reconstructs the original sentence, correctly preserving names, dates, numeric ranges, and grammatical structure, closely matching the reference.}
  \label{fig:brl2kr-ours}
\end{figure*}

\begin{figure*}[t]
  \centering
  \includegraphics[width=\linewidth]{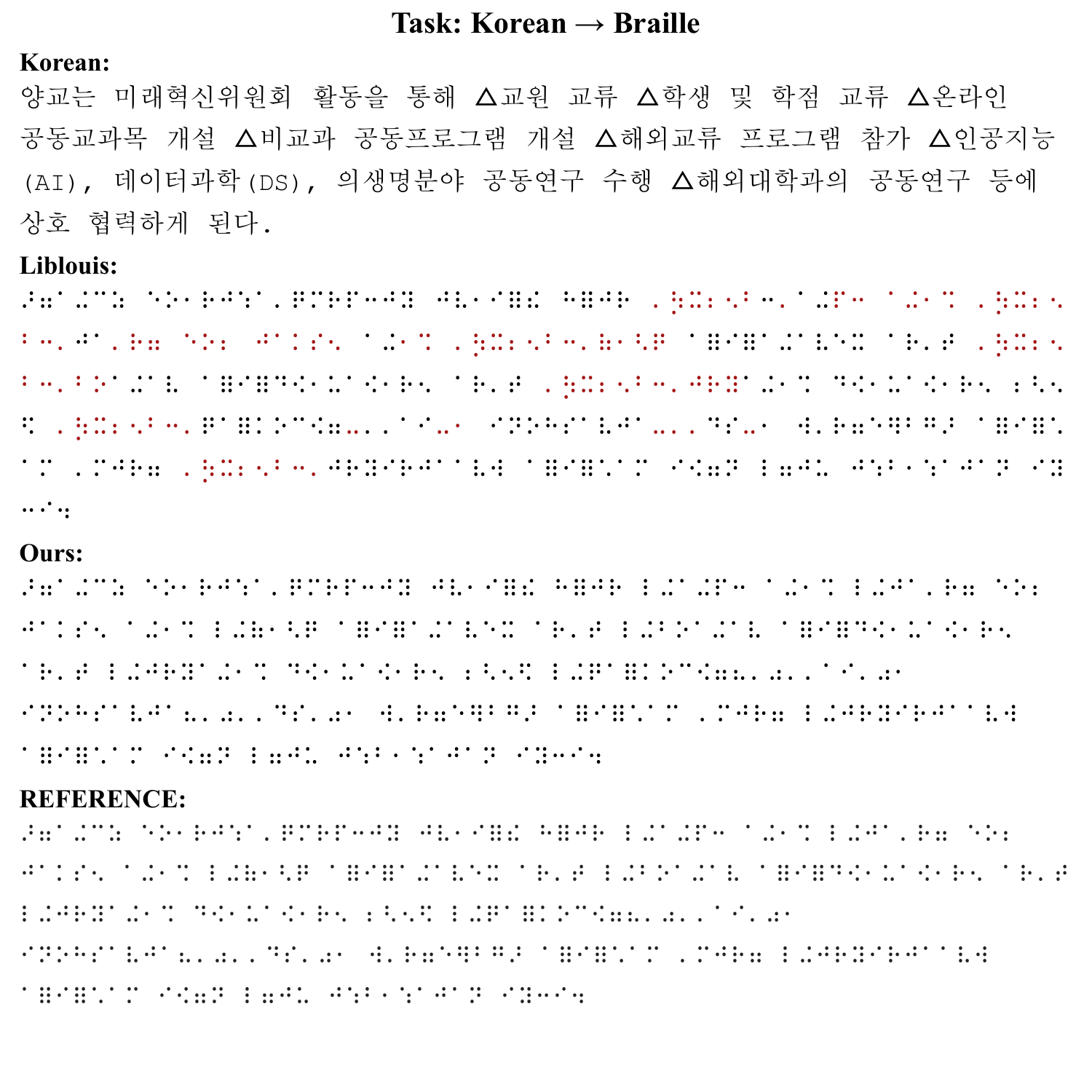}
  \caption{Example outputs for the Korean-to--Braille task for the rule-based Liblouis system and the BT5 model. Incorrect or mismatched segments are highlighted in red. Liblouis produces structurally inconsistent Braille sequences due to rigid rule application, whereas BT5 closely matches the reference output.}
  \label{fig:kr2brl-ours}
\end{figure*}

\end{document}